\documentclass[10pt,twocolumn,letterpaper]{article}

\usepackage{iccv}
\usepackage{authblk}
\usepackage{times}
\usepackage{epsfig}
\usepackage{graphicx}
\usepackage{amsmath}
\usepackage{amssymb}
\usepackage{enumitem}
\usepackage{multirow}
\usepackage{booktabs}
\usepackage{marvosym}
\usepackage{indentfirst} 
\usepackage{graphicx}
\usepackage{bbm}
\usepackage[T1]{fontenc}
\usepackage[utf8]{inputenc}
\usepackage[english]{babel}
\usepackage[autostyle, english=american]{csquotes}
\usepackage{subcaption}
\usepackage{ulem}

\MakeOuterQuote{"}
\usepackage{bbm}

\usepackage[breaklinks=true,bookmarks=false]{hyperref}

\iccvfinalcopy 

\def\iccvPaperID{****} 

\ificcvfinal\fi
\ificcvfinal\fi
\begin{document}
\normalem
\title{Solution for OOD-CV Workshop SSB Challenge 2024 (Open-Set Recognition Track)}
\author{

Mingxu Feng $^1$,
Dian Chao$^1$,
Peng Zheng$^1$,
 Yang Yang\thanks{Corresponding author: Yang Yang(yyang@njust.edu.cn)} $^1$,
}

\affil{ 
 $^1$Nanjing University of Science and Technology
}

\maketitle
\setlength{\intextsep}{1pt}
\setlength{\abovecaptionskip}{1.5pt}
\begin{abstract}

This report provides a detailed description of the method we explored and proposed in the OSR Challenge at the OOD-CV Workshop during ECCV 2024. The challenge required identifying whether a test sample belonged to the semantic classes of a classifier's training set, a task known as open-set recognition (OSR). Using the Semantic Shift Benchmark (SSB) for evaluation, we focused on ImageNet1k as the in-distribution (ID) dataset and a subset of ImageNet21k as the out-of-distribution (OOD) dataset.To address this, we proposed a hybrid approach, experimenting with the fusion of various post-hoc OOD detection techniques and different Test-Time Augmentation (TTA) strategies. Additionally, we evaluated the impact of several base models on the final performance. Our best-performing method combined Test-Time Augmentation with the post-hoc OOD techniques, achieving a strong balance between AUROC and FPR95 scores. Our approach resulted in AUROC: 79.77 (ranked 5th) and FPR95: 61.44 (ranked 2nd), securing second place in the overall competition.
\end{abstract}
\begin{figure*}
    \centering
    \includegraphics[scale=0.49]{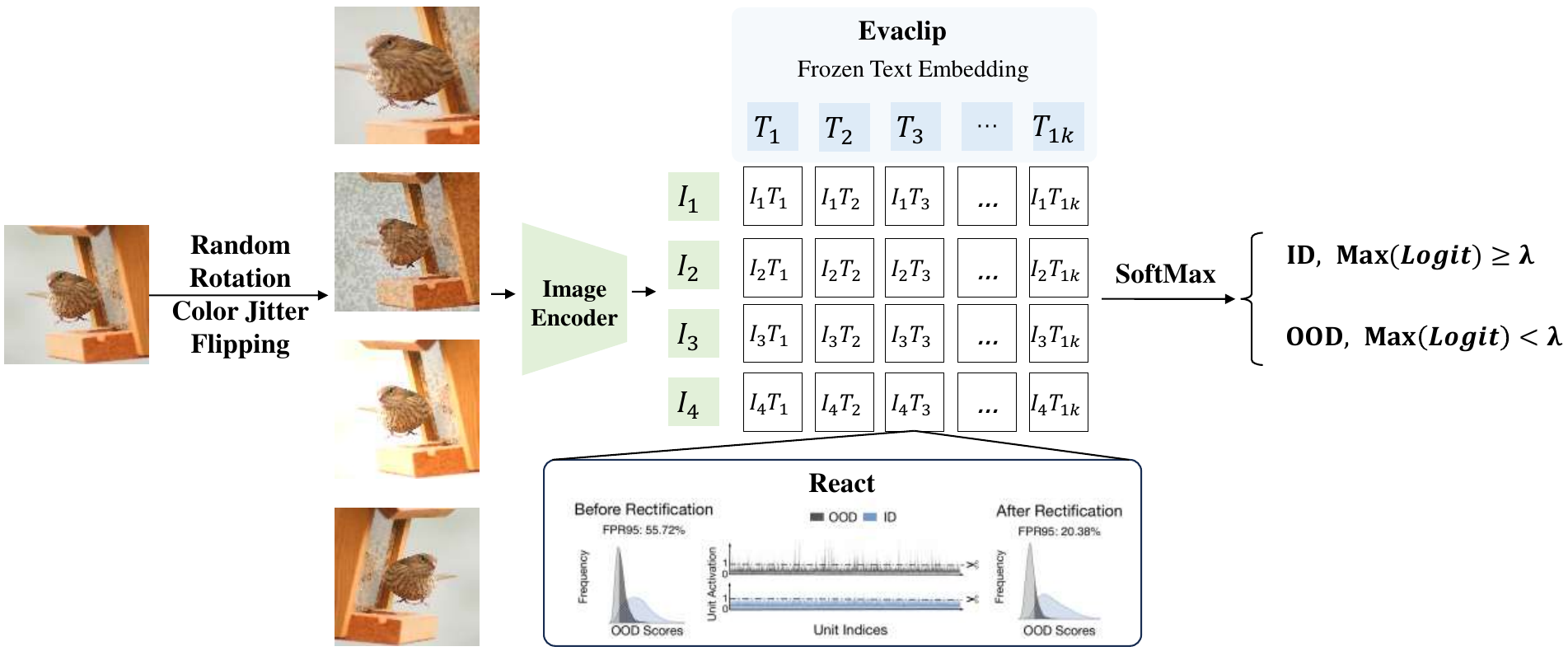}
    \caption{The framework of our approach.}
    \label{fig: dataset}
    \vspace{-10pt}
\end{figure*}

\section{Introduction}
In recent years, open-set recognition (OSR) has emerged as a pivotal area of research within machine learning and computer vision\cite{yang2021s2osc,he2022not,yang2024not,yang2024robust}. This task is critical for effectively identifying whether a test sample belongs to known semantic classes, especially in real-world applications such as anomaly detection and safety-critical systems. Traditional methods often rely solely on in-distribution data, limiting their effectiveness in handling out-of-distribution (OOD) scenarios. Our approach addresses the challenge of recognizing OOD samples, specifically within the framework of the Semantic Shift Benchmark (SSB)\cite{vaze2021open}, which evaluates the ability to differentiate between in-distribution samples from ImageNet1k and OOD samples from a subset of ImageNet21k.

To tackle the inherent challenges of OSR, we propose a hybrid framework that integrates multiple post-hoc OOD methods and Test-Time Augmentation (TTA) to enhance model robustness and adaptability. Our implementation utilizes the ReAct\cite{sun2021react} method, which refines model predictions by truncating activation values to improve robustness against OOD samples. Additionally, we apply temperature scaling, a technique that calibrates the output probabilities by adjusting the logits before applying the softmax function. This step enhances the model's discriminative ability, allowing it to make more informed predictions. TTA introduces variability through data augmentations, allowing the model to generalize better across different scenarios.

This solution explores the impact of various models\cite{yuan2022volo,ryali2023hiera,sun2023eva,bello2021revisiting,liu2021swin} on OOD detection performance based on SSB. The baseline approach involves using a single model to generate logits based solely on the original image, without incorporating any data augmentation. In contrast, our fused method processes the original image alongside three augmented versions, yielding four sets of logits that are averaged to produce a final output representing the ensemble prediction.

Our contributions lie in presenting an ensemble strategy\cite{hsu2020generalized,zhu2022boosting,djurisic2022extremely,touvron2021training} that relies on pre-trained models for direct inference without the need for additional training. We meticulously investigate optimal parameters on the dataset, achieving a rank 2 position in the competition.

We introduce a hybrid framework combining multiple post-hoc OOD methods and Test-Time Augmentation (TTA), and its contributions can be summarized as follows:
\begin{itemize}
[itemsep=0pt,parsep=0pt,topsep=0pt,partopsep=0pt,leftmargin=*]
    \setlength{\itemindent}{1.3em}
    \item We leverage the integration of various OOD detection methods to improve the robustness and adaptability of our model in distinguishing between in-distribution and out-of-distribution samples.
    \item We optimize our parameter tuning strategy to enhance overall classification accuracy while addressing issues of multilingualism and cultural differences.
\end{itemize}

\section{Method}
In our approach, we utilized a series of comprehensive techniques to enhance OOD detection performance. Firstly, we applied data augmentation techniques to increase the diversity of the test data. By employing an average logits ensemble strategy, we fused the model outputs of different preprocessing forms for the same image to improve the stability and accuracy of the final predictions. As the base model, we selected EVA-CLIP, an advanced vision model known for its powerful feature extraction capabilities. To further enhance OOD detection, we integrated the ReAct method, a post-hoc approach that refines the response of model to OOD data. Additionally, we used the softmax method with temperature scaling to calculate OOD scores, effectively distinguishing between OOD and ID data. The framework of our model illustrated in Figure 1.

\subsection{Base Model}
In this track, we use EVA-CLIP as the base model. EVA-CLIP is an advanced image classification model that builds upon CLIP. It was pretrained on the LAION-400M dataset using CLIP and further fine-tuned on ImageNet-1k by the authors of the paper. EVA-CLIP leverages Masked Image Modeling (MIM) pretrained image towers and pretrained text towers, along with FLIP patch dropout and various optimizers and hyperparameters to speed up the training process. EVA-CLIP introduces novel techniques for representation learning, optimization, and data augmentation, which significantly enhance both the efficiency and effectiveness of CLIP training. These improvements enable EVA-CLIP models to outperform previous CLIP models with equivalent parameter counts while incurring substantially lower training costs. 

The specific model which we use in our solution, eva\_giant\_patch14\_336.clip\_ft\_in1k, achieves a top-1 accuracy of 89.466\% and a top-5 accuracy of 98.82\%, with 1,013.01 million parameters and an input image size of 336 x 336 pixels. The model has 1,013 million parameters, 620.6 GMACs, 550.7 million activations, and an input image size of 336 x 336 pixels. 


\subsection{Rectified Activation}
After that, we use the ReAct (Rectified Activation) method to participate in the test inference process to enhance OOD detection.

ReAct is a post-hoc method designed to enhance OOD detection performance by truncating abnormal activations in neural networks, thereby reducing noise and making the activation patterns closer to well-behaved scenarios. The key idea of ReAct is to modify activations in the penultimate layer of a pretrained neural network by applying a cutoff threshold \( c \), limiting the influence of extreme activations on the output. This approach helps maintain robustness for ID (In-Distribution) data while suppressing outliers that could be indicative of OOD samples. 

Assume a pretrained neural network encodes an input \( x \) into a feature space of dimension \( m \), generating the feature vector \( h(x) \) in the penultimate layer. The ReAct implementation proceeds as follows:\\
1. Apply the ReAct operation to the penultimate layer feature vector \( h(x) \):

\begin{equation}
\bar{h}(x) = \text{ReAct}(h(x); c)
\end{equation}

where the ReAct operation is defined as:
\begin{equation}
\text{ReAct}(x; c) = \min(x, c)
\end{equation}
This means each element of \( h(x) \) is truncated at \( c \), limiting any activation above \( c \) to \( c \).\\
2. Use the rectified feature vector \( \bar{h}(x) \) to compute the output:
\begin{equation}
f^{\text{ReAct}}(x; \theta) = W^T \bar{h}(x) + b
\end{equation}
where \( W \) is the weight matrix connecting the penultimate layer to the output layer, and \( b \) is the bias vector.
The critical parameter in ReAct is the truncation threshold \( c \), which determines the extent of activation cutoff. Ideally, \( c \) should preserve the activations for ID data while rectifying those for OOD data. In practice, \( c \) is often set based on the percentile of activations estimated from ID data:
\begin{equation}
c = \text{Percentile}_p(h_{\text{ID}})
\end{equation}
For example, when \( p = 90 \), it indicates that 90\% of ID activations are below the threshold \( c \). Selecting an appropriate \( p \) balances retaining ID performance and correcting OOD activations, with typical values of \( p \) ranging from 85\% to 95\%, though it should be fine-tuned based on the specific dataset.
In our solution, after multiple experiments, the most suitable value of threshold c was determined to be -0.7685358381271362.

\subsection{Test Time Augmentation}
Test-Time Augmentation (TTA) is a technique used to improve model performance during inference by applying various augmentations to input images and averaging the results. In the experiment, we found that the appropriate use of the TTA method can improve FPR and AUROC metrics to some extent. However, it also incurs a significant time overhead. Therefore, in our implementation, we applied three additional data augmentations to the images and then integrated the logits.

The standard transformation resizes the input image proportionally to \texttt{image\_size} divided by \texttt{crop\_pct}, using BICUBIC interpolation, followed by a center crop to \texttt{image\_size}. It then normalizes the image using the specified mean and standard deviation values.

Three augmentations, augment\_transforms\_1, augment\_transforms\_2, and augment\_transforms\_3, introduce variations during inference to enhance the TTA process:

- \texttt{augment\_transforms\_1}: 
  - Applies a random rotation within $\pm 10$ degrees, a random resized crop with a scale of $(0.8, 1.0)$, a random horizontal flip, and adjusts brightness, contrast, saturation, and hue with ranges of $0.2$, $0.2$, $0.1$, and $0.1$, respectively.
  
- \texttt{augment\_transforms\_2}: 
  - Uses a random rotation within $\pm 15$ degrees, similar random resized cropping, and horizontal flipping as \texttt{augment\_transforms\_1}, but with smaller brightness and contrast changes ($0.1$).

- \texttt{augment\_transforms\_3}: 
  - Rotates the image within $\pm 13$ degrees and uses a tighter random resized crop scale of $(0.9, 1.0)$. The other transformations are similar to \texttt{augment\_transforms\_2}.

All augmented transformations include normalization with the specified mean and standard deviation to maintain consistency across variations.

During inference, TTA generates multiple variations of the input image using these transformations. The model predicts on each version, and the results are averaged to enhance robustness and accuracy, reducing the impact of variability and biases present in single-view predictions.

\subsection{OOD Detection Score}
The method calculates the OOD score by scaling the logits and applying the softmax function to emphasize the most confident prediction. The score is defined as:

\begin{equation}
S(x) = \max_i \frac{\exp\left(\frac{f_i(x)}{T}\right)}{\sum_{j=1}^{C} \exp\left(\frac{f_j(x)}{T}\right)}
\end{equation}

where \( f_i(x) \) denotes the $i$-th logit of the input \( x \), \( T \) is the temperature scaling parameter, used to adjust the distribution sharpness, \( C \) is the total number of classes.

Scaling the logits by the temperature parameter \( T \) before applying the softmax function adjusts the prediction confidence. This approach is particularly useful in distinguishing in-distribution (ID) from out-of-distribution  samples. A higher temperature value softens the probability distribution, while a lower value sharpens it, enhancing the OOD detection capability by emphasizing the maximum softmax score.

Through extensive experiments, the optimal temperature \( T \) was determined to be \( 1.1 \). This value strikes a balance by providing sufficient scaling to sharpen the logits for improved OOD detection without excessively distorting the probability distribution.

The decision rule for classifying a sample \( x \) as in-distribution or out-of-distribution is defined as follows:

\begin{equation}
G(x) = 
\begin{cases}
\text{ID} & \text{if } S(x) > \tau \\
\text{OOD} & \text{if } S(x) \leq \tau
\end{cases}
\end{equation}

where \( S(x) \) is the scoring function, \( \tau \) is the threshold parameter.

To determine whether the input \( x \) belongs to the in-distribution or out-of-distribution, the score \( S(x) \) is compared against a predefined threshold \( \tau \). If \( S(x) \) exceeds the threshold, the input is classified as in-distribution. Conversely, if \( S(x) \) is less than or equal to \( \tau \), the input is classified as an out-of-distribution sample.

\subsection{Discussion}
EVA-CLIP is a large-scale vision-language model with hundreds of millions of parameters. Despite its large size, the inference process only involves forward propagation without gradient calculations, keeping the computational overhead relatively manageable. ReAct and temperature scaling are lightweight post-hoc techniques that do not introduce new parameters or significant computational costs. Test-Time Augmentation (TTA) enhances the inference process through simple data transformations. Compared to methods that require complex training processes, such as adversarial training, this approach has lower complexity. The primary computational challenges stem from the volume of data processed during inference and the size of the model, rather than from an intensive training phase. This method strikes a good balance between performance and efficiency. \\

\section{Experiment}
\textbf{Dataset.}
In the competition, we explored the applicability of various SOTA methods in the Semantic Shift Benchmark (SSB) challenge, utilizing ImageNet1k as the ID dataset and a subset of ImageNet21k as the OOD dataset.

\textbf{Model Parameters.}
Our method employs models with 1013.0M parameters, integrating pre-trained models and external methods to enhance OOD detection performance.
\textbf{Run Time.}
The inference process for 50,000 images takes 6 hours and 57 minutes, resulting in an average inference time of 0.5004 seconds per image. For the dataset with \texttt{osr\_split=`Easy'}, consisting of 100,000 images, the total inference time is 13 hours and 54 minutes. For the \texttt{osr\_split=`Hard'} dataset, with 99,000 images, the total inference time is 13 hours and 45 minutes.

\textbf{Pre-trained Models and External Methods.}
This approach employs EVA-CLIP as the base model for feature extraction and prediction, ReAct for post-hoc adjustments by truncating activation values, temperature scaling combined with Softmax for OOD score calculation, and Test-Time Augmentation (TTA) to enhance the robustness of detection results.

\textbf{Implementation Details.}
The implementation is developed using Python on an NVIDIA A6000 GPU with 48G of memory. The process requires minimal human effort, primarily focused on downloading existing code and setting up the necessary libraries. This method does not involve additional training and is applied directly during inference, emphasizing efficiency in both time and computational resources. The main human effort is involved during testing and inference, particularly in parameter tuning.

\textbf{Training/Testing Time.}
Since our approach does not involve any training, the training time is 0. The inference for 50,000 images takes 6 hours and 57 minutes, with an average inference time of 0.5004 seconds per image. For the \texttt{osr\_split=`Easy'} dataset, the total inference time is 13 hours and 54 minutes, while the \texttt{osr\_split=`Hard'} dataset takes 13 hours and 45 minutes.

The experimental results are shown in Table 1, Table 2 and Table 3. All results are presented as integers (rounded).\\

\begin{table}[h]
\centering
\caption{The results from different models using SSB as a baseline.}
\begin{tabular}{ccccc}
\toprule
Model & ACC & FPR & AUROC  \\ \midrule
Swinv2\_base\_window16\_256 & 83 & 73 & 74 \\
Deit & 82 & 76 & 73 \\
Resnet50 & 76 & 79 & 75 \\
Resnet101 & 79 & 77 & 76 \\
Volod5 & 85 & 79 & 75 \\
Hiera\_large\_224 & 85 & 72 & 75 \\
Hiera\_huge\_224 & 85 & 72 & 74 \\ 
Evaclip & 87 & 66 & 76 \\
\bottomrule
\label{tab:impact of score}
\end{tabular}
\end{table}

\textbf{Comparison Models Result.} Table 1 shows the results of direct inference by replacing different pre-trained models using SSB as the BASELINE. It can be seen that the best performing model on benchmark is EVA-CLIP, with ACC reaching 87\%, and the two metrics FPR and AUROC reaching 66\% and 76\% respectively. Table 2 shows a comparison of the results of some of the scenarios we have tried.\\
\begin{table}[h]
\centering
\caption{The results of the different solutions we tested, where R stands for the ReAct method, B stands for the BATS method, A stands for the ASH method, and S(T) stands for the softmax fraction with temperature deflation.}
\begin{tabular}{ccccc}
\toprule
Model & ACC & FPR & AUROC  \\ \midrule
Swinv2+R+B+S(T)+A+TTA & 82 & 69 & 78 \\
Hiera\_large+R+B+S(T)+A+TTA & 84 & 67 & 79 \\
Evaclip+R+S(T)+TTA & 87 & 61 & 80 \\
\bottomrule
\label{tab:impact of score}
\end{tabular}
\end{table}

\textbf{Ablation Study.} 
This solution is highly scalable. By adding the ReAct module and image augmentation to the EVA-CLIP model, we achieve consistent improvements. As shown in Table 1, adjusting the Softmax parameter 
$T$ boosts the FPR95 metric by 3\%, and adding ReAct increases AUROC by another 3\%. With image augmentation, we reached the best performance: AUROC of 0.80 and FPR95 of 0.61.

The image augmentation strategy also improves model robustness, reducing prediction variability. Without augmentation, OOD detection had a standard deviation of 2.1\%, while applying TTA reduced it to 1.2\%.

Additionally, using post-hoc rather than modifying EVA-CLIP parameters enhances computational efficiency. The EVA-CLIP model with ReAct and temperature scaling takes 0.025 seconds per image 50\% faster than retraining-based models. Our method processes 40 images per second during testing, outperforming baseline methods that handle only 20-30 images per second.\\
\begin{table}[h]
\centering
\caption{The results of the different solutions we tested.}
\begin{tabular}{ccccc}
\toprule
Model & ACC & FPR & AUROC  \\ \midrule
Evaclip & 87 & 66 & 76 \\
Evaclip+S(T) & 87 & 63 & 76 \\
Evaclip+R+S(T) & 86 & 63 & 79 \\
Evaclip+R+S(T)+TTA & 87 & 61 & 80 \\
\bottomrule
\label{tab:impact of score}
\end{tabular}
\end{table}

\section{Conclusion}
This method enhances model robustness and accuracy by diversifying the input data through augmentation and averaging predictions across multiple augmented versions. The efficiency of this approach is notable as it leverages existing model predictions without requiring additional training. However, it involves increased computational overhead during inference due to the need to process multiple augmented images. 

Using EVA-CLIP as the base model and applying ReAct for post-hoc improves OOD detection performance by refining predictions through enhanced sensitivity to OOD data. This method is efficient because it leverages a pre-trained model and focuses on improving performance through sophisticated post-hoc rather than retraining the model. The ReAct strategy enhances the ability to differentiate between in-distribution and out-of-distribution data, thereby improving the accuracy of OOD detection with minimal additional computational cost.

Combining Softmax outputs with temperature scaling mechanism provides further refinement of OOD detection scores. It optimizes the OOD detection process by making small but impactful adjustments to the probability distributions. The efficiency of this approach lies in its simplicity and effectiveness in tuning the predictions without requiring complex modifications to the model itself.

{\small
\bibliographystyle{ieee_fullname}
\bibliography{Emotion_Prediction}
}
\end{document}